\documentclass[10pt,twocolumn,letterpaper]{article}

\usepackage[pagenumbers]{cvpr}      

\usepackage{graphicx}
\usepackage{amsmath}
\usepackage{amssymb}
\usepackage{booktabs}

\usepackage{xcolor} 
\newcommand{\fred}[1]{\textcolor{black}{#1}}  
\newcommand{\muna}[1]{\textcolor{black}{#1}}

\usepackage{hyperref}
\hypersetup{pagebackref,breaklinks,colorlinks}

\usepackage[capitalize]{cleveref}
\crefname{section}{Sec.}{Secs.}
\Crefname{section}{Section}{Sections}
\Crefname{table}{Table}{Tables}
\crefname{table}{Tab.}{Tabs.}

\def\cvprPaperID{9116} 
\def\confName{CVPR}
\def\confYear{2022}

\begin{document}

\title{Distillation of Human-Object Interaction Contexts for Action Recognition}

\author{Muna Almushyti\\
Durham University, UK\\
{\tt\small muna.i.almushyti@durham.ac.uk}
\and
 Frederick W. Li\\
Durham University, UK\\
{\tt\small frederick.li@durham.ac.uk}
}
\maketitle

\begin{abstract}
   \fred{Modeling spatial-temporal relations is imperative for recognizing human actions, especially when a human is interacting with objects, while multiple objects appear around the human differently over time.
   Most existing action recognition models focus on learning overall visual cues of a scene but disregard informative fine-grained features, which can be captured by learning human-object relationships and interactions.
   In this paper, we learn human-object relationships by exploiting the interaction of their local and global contexts. We hence propose the Global-Local Interaction Distillation Network (GLIDN), learning human and object interactions through space and time via knowledge distillation for fine-grained scene understanding. GLIDN encodes humans and objects into graph nodes and learns local and global relations via graph attention network \cite{velivckovic2017graph}. The local context graphs learn the relation between humans and objects at a frame level by capturing their co-occurrence at a specific time step.
   The global relation graph is constructed based on the video-level of human and object interactions, identifying their long-term relations throughout a video sequence.
   More importantly, we investigate how knowledge from these graphs can be distilled to their counterparts for improving human-object interaction (HOI) recognition. We evaluate our model by conducting comprehensive experiments on two datasets including Charades \cite{sigurdsson2016hollywood} and CAD-120 \cite{koppula2013learning} datasets. We have achieved better results than the baselines and counterpart approaches.}

\end{abstract}
\begin{figure}[t]
  \centering
   \includegraphics[width=1.2\linewidth]{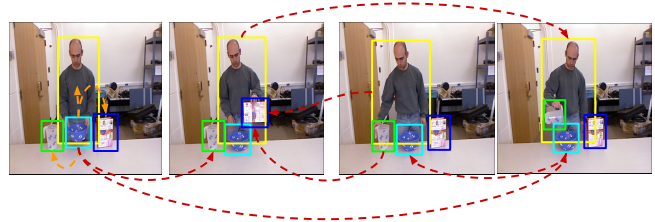}
   \caption{\fred{Local context (orange dash arrows) can provide information about interactions among human and objects at a specific time. Global context (red dash arrows) provides a view of HOIs over time.}}
   \label{fig:general_idea}
\end{figure}
\section{Introduction}
Human action recognition tasks typically involve interaction with objects. Such tasks are challenging even for deep learning methods especially under complex scenarios. A human can interact with the same object but performing different actions. For example, a human can hold a laptop and can put it somewhere. These two actions, “hold” and “put”, are different but they involve the same object. In addition, a variety types of objects afforded to same action (e.g., refrigerators and doors can be involved in the same interactions including open and close) needs to be considered \cite{xu2019learning}. Moreover, the existence of different objects around a human could affect model predictions. For example, if a human is drinking a coffee and there is a book nearby, a model may inaccurately predict that the human is both reading and drinking. Furthermore, during a video sequence, the states of humans and objects change over time, such as a human can hold an object and release it at any time step, followed by interacting with another object which makes identifying correct interactions very challenging. Hence, identifying humans and objects at each time steps and learning their relations can help understand a scene. This implies learning objects that are closely located for identifying interactions. The transition of human and object states over time also offers crucial cues for understanding what a human is performing. \fred{Consequently, it is important to capture contextual information about interactions both at a specific time and throughout a video, making HOI recognition success.}

Although modelling HOIs has been broadly studied in images  \cite{chao2018learning,gkioxari2018detecting,bansal2019detecting,xu2019interact}, it has received less consideration in videos. Even deep learning methods have been developed for recognizing human actions in videos, most of them, including 
Covnet \cite{simonyan2014two}, recurrent neural networks (RNNs) \cite{donahue2015long,li2017temporal} and 3D convolution models \cite{carreira2017quo,tran2018closer}, 
\fred{only take individual frame-wise information as inputs (coarse-grained) without explicitly modeling (fine-grained) human-object relations across a video sequence. Hence, such methods failed to capture useful global context cues, i.e. long-term human object dependency, for assisting action recognition.}

Recent works \cite{wang2018videos,herzig2019spatio,tan2019object, materzynska2020something,baradel2018object} have proposed to model human-object relations by performing spatio-temporal reasoning through multi-head attention mechanism for recognizing actions in videos. \fred{As they capture more context cues to reason HOIs, they have achieved promising results over baselines that do not consider human-object relations.}

\fred{In this work, we propose to capture human-object relations from their local and global views as well as transferring knowledge between these views. The local view captures human-object relations at a specific time, e.g., spatial relation. The global view encodes human-object relations over time, e.g., temporal relation, to capture long-term human-object relations.} 
\muna{The design of the network for global and local views is flexible. Inspired by the success of graph attention networks (GAT) \cite{velivckovic2017graph} in different tasks including person re-identification\cite{yang2020spatial}, action recognition \cite{yan2018spatial,wang2018videos,lu2019gaim} and video question answering \cite{huang2020location}, we exploit them to construct our two contextual views modules.}

Since the global context of an interaction offers complementary information to the local contexts of such interaction and vice versa, previous works combined different types of context features via concatenation \cite{materzynska2020something} or summation \cite{wang2018videos}, or even considered the global features as an extra node in the graph \cite{ghosh2020stacked}.
Inspired by \cite{pan2020spatio} and instead of learning these contexts via features level which are prone to noise, we propose to apply knowledge distillation, transferring knowledge about interactions from global to local views, and vice versa. \fred{We therefore exploit teacher-student network design, investigating which of the proposed contextual views can form a better teacher, which offers richer HOI information, to guide the student network for improving action recognition performance.}

To the best of our knowledge, we are the first to investigate knowledge distillations between HOI graphs for action recognition in videos.
Our main contributions are:
\begin{itemize}
\item Proposing a novel teacher-student network based on graphs neural networks to learn spatial and temporal interrelations between humans and objects in a video from two different contextual views. Hence,  long-term and non-local dependency between human and objects across video frames can be captured.

\item Investigating how structural knowledge from the teacher contextual view of interactions can be obtained, and distilling it to the student view of interactions to improve action recognition performance.

\item
Evaluating our model on Charades and CAD-120\cite{koppula2013learning} datasets \cite{sigurdsson2016hollywood} and conducting comprehensive experiments in transferring knowledge between local (e.g., Spatial) and global (e.g., Temporal) contexts of human-object interactions. {Our teacher-student design is effective to distill knowledge to local context graphs from global context. We also observe that the student network outperforms its teacher by exploiting both global and local contexts of an interaction.}
\end{itemize}

\section{Related Work}
\paragraph{Action recognition models in videos.} The simple models for action recognition can be done by extracting frame features through CNNs followed by pooling via averaging, or followed by RNNs to model the sequence of frames for predicting actions in videos \cite{donahue2015long,yue2015beyond}. Recently,  space-time models are proposed, such as 3D convolutions. They add an extra time dimension to kernels in order to extract spatio-temporal features from videos \cite{ji20123d,tran2015learning,varol2017long,feichtenhofer2019slowfast}.
Likewise, I3D model \cite{carreira2017quo} has been introduced by inflating pretrained 2D convolution kernels to 3D for extracting space-time features from video clips. 
There are related methods focusing on long-term dependency as in \cite{zhou2018temporal} where the temporal relations between frames at different time scales are modeled via multilayer perceptrons. Also, non-local relations between pixels in space and time are studied for recognizing actions in videos.

More Recently, transformer-based frameworks such as \cite{neimark2021video} are proposed for recognizing actions in videos where the transformer is used to get discriminative features from each frame and then being aggregated via attention. Transformers is also used for action recognition networks purely without utilizing convolutions \cite{arnab2021vivit}. In addition, beside the appearance features that can be extracted from RGB images, optical flow and depth data are used to enhance human action recognition in videos \cite{simonyan2014two,Cheng_2020_CVPR,chengdecoupling,si2018skeleton}.

All the above mentioned efforts focus on whole video features (coarse-grained) rather than on key cues of an action such as inter-objects or inter-human relations that our work considers. Also, our method only focuses on visual information from videos to model HOIs for action recognition.

\paragraph{Spatio-temporal reasoning for action recognition.} 
Spatio-temporal reasoning involves detecting humans and objects and modeling their relations in order to capture contextual information that helps classify an action. In \cite{wu2019long}, the relation between objects at specific time and the objects from adjacent frames at specific window is learned via Feature Bank Operator (FBO), such as non-local, to capture long-term context in videos. Moreover, inspired by the success of recurrent neural networks (RNNs) in modeling sequence data, such as Long Short-Term Memory (LSTM), they have also been used for spatio-temporal reasoning over objects in videos \cite{baradel2018object}.

Space and time graphs have been proposed in \cite{wang2018videos}, where object context relations during time is captured and objects in adjacent frames are connected based on their intersection over unions (IOU). A relation network is proposed focusing on the relation between actors and video-level features for identifying actions\cite{sun2018actor}. \fred{To capture high order object interactions, attention mechanism is applied over objects at each frame followed by a LSTM process.}

Furthermore, in \cite{tomei2021video}, graph attention is used to model the relations between human and objects, considering their spatial distance in each clip. Transformers are also used in learning visual relations between the features of humans located in the centre clip, which is considered as a query, and the features from the whole clip in order to learn the context of the action by using the properties of self-attention in the transformer \cite{girdhar2019video}. 
Our work propose to use two different contexts of human and object relations, capturing different cues of an interaction that helps recognize actions. Inspired by \cite{velivckovic2017graph,yang2020spatial}, we choose graph attention network as a base network for learning such interactions.

\paragraph{Knowledge distillation (KD).}
Distilling knowledge has been proposed to transfer knowledge learned from ensemble of classifiers or large network into a small network \cite{hinton2015distilling}. This implies compressing complex networks without losing their performance \cite{thoker2019cross}. It can be done by minimizing the loss between small network (student) predictions and the large network's soften labels (teacher). Recently, the concept of KD is extended and combined with privileged information \cite {vapnik2009new}, where additional information is available only during training time to form a generalized distillation \cite{lopez2015unifying}. For action recognition task, the knowledge is distilled between multiple modalities (e.g., skeleton, RGB), which can be considered as privilege information and not all of them are available during inference \cite{luo2018graph,garcia2021distillation}. Moreover, the concept of KD is employed in different directions,  such as defencing against adversarial attacks \cite{papernot2016distillation}, classifying unlabeled data via unifying diverse classifiers \cite{vongkulbhisal2019unifying}. Inspired by these directions, we extend it to HOI recognition in videos, allowing knowledge transfer between global and local contextual views of interactions via KD.
\begin{figure*}
  \centering
   \includegraphics[width=1.0\linewidth]{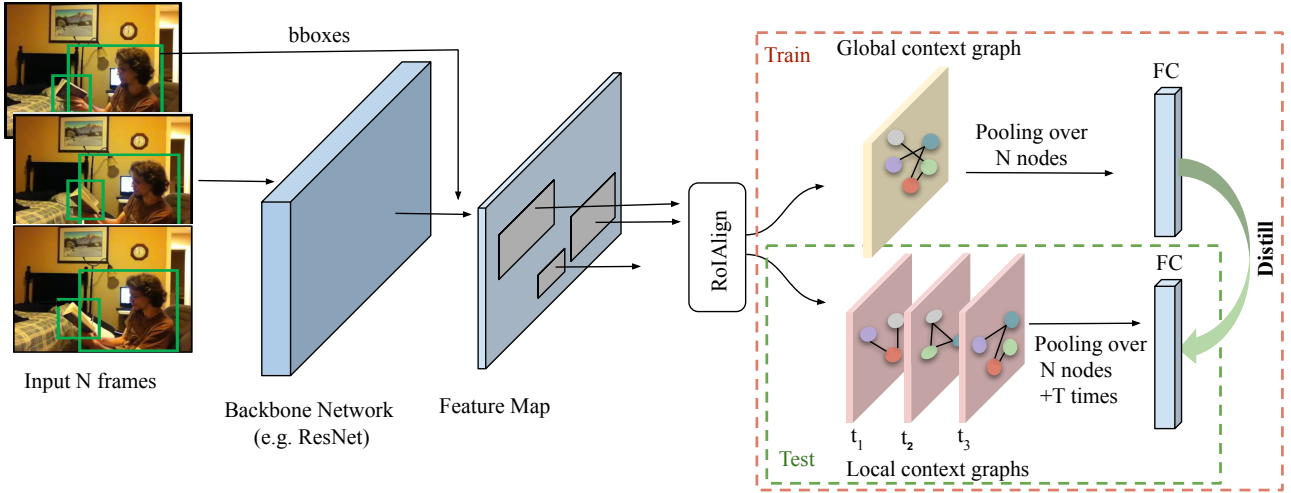}

   \caption{Overview of our proposed GLIDN network.}
   \label{fig:overview}
\end{figure*}

\section{Global-Local Interaction Distillation Network (GLIDN)}

\subsection{Network overview}
Figure \ref{fig:overview} shows the architecture of our GLIDN. It takes video frames and the bounding boxes of human and objects at each frame as inputs. Frame features (e.g., appearance features) are then extracted by a convolutional neural network, such as ResNet \cite{he2016deep}. 
RoIAlign \cite{he2017mask} is then applied to extract features of each human and object boxes from the backbone feature map. The bounding boxes are generated via Region Proposal Network \cite{ren2015faster} if they are not available in the dataset. These extracted region features are used as the initial features of graph nodes in both the global and local contextual views. The human-objects relations from the teacher view are distilled into the student context representation by aligning logits from the two contextual views.

\subsection{Global and Local Context Graphs}

As mention earlier, we utilize graph attention network (GAT)\cite{velivckovic2017graph} as our graph networks to learn the relations between human and objects from different contextual views.

The global context graph is constructed to learn the relation between each entity (e.g., human or object) and all other entities in a video. The graph is constructed based on learned adjacency matrix between humans and objects over time in a video as in \cite{wang2018videos}. Hence, the interaction score between two nodes in GAT can be computed as:
\begin{equation}
 \alpha_{i,j} =\sigma(a [W_{o}(x_{i})| W_{o}(x_{j})])
 \label{eq:1}
\end{equation}
where $W_{o}$ is a learnable transformation which is shared between nodes in a video. $a$ is a weight matrix projecting the concatenated features to a scalar that reflects attention coefficients between two nodes (e.g., humans or objects). "$|$" indicates concatenation. In this global context graph, coefficients represent the learned interaction score between humans and objects. In other words, $\alpha_{i,j}$ is a scalar that represents the relation between two nodes $i$ and $j$ in the adjacency matrix $A$, which is of the size $N\times N$ where $N$ is the number of humans and objects that appeared in the video. $\sigma$ is a nonlinearity function such as LeakyReLU. Later, $\alpha$ is normalized across all other nodes within the video with respect to node $i$ via softmax. Thus, the updated node features via GAT can be formulated as:
\begin{equation}
 x_{i} = \sum_{j\in N} \alpha_{i,j}  W_{o}  x_{j}
 \label{eq:2}
\end{equation}
Through this graph, long-term dependency of HOIs in a video can be captured since each object is attended to all other objects over the video at different time frames.

On the other hand, in the local context, there are T number of graphs, where T indicates the number of frames in the video. Through these local graphs, \fred{besides relations induced by closely located humans and objects}, non-local dependency relations between human and objects in a video frame can also be captured. Non-local means when objects and humans are distant from each other within a frame. Hence, each node captures local contextual information via learning relation with other nodes (e.g., human or objects) within the same frame 
\fred{regardless they are spatially close to or distant from each other. Local context is therefore learned from various interactions in which humans / objects attend to others in the same frame.}

\fred{Through these graphs, the relation between humans and objects is learned even though they are not nearby in space and time which extensively learning various human-object, object-object and human-human relations both within individual frames and throughout a video.}

\subsection{Global and Local Context Distillation}
In order to have an informative representation of HOIs, considering global and context relations between human and objects, 
\fred{features from both contextual views should be fully utilized. This may not be simply done by combining features from the two contexts, despite it is a standard way for gathering information from different sources or views. In contrast, we adapt a teacher-student framework to utilize global and local context of HOIs through knowledge distillation. To implement such a knowledge transfer, we incorporate soft labels from the teacher context graph network to guide the student context graph network during training, where these soft targets are probability distributions from the logits in the teacher network.}

Our experiments two different distillation losses depending on the nature of a dataset. For CAD-120 dataset, we minimize the KL divergence between soften labels of teacher and student as in \cite{pan2020spatio,bian2021structural}. For Charades, we use $l_{2}$ loss as distillation loss to meet the property of multi-label classification task. Hence, the $l_{2}$ distillation loss can be formulated as \cite{liu2018multi}:
\begin{equation}
\begin{aligned}
 L_{Distill} =\frac{1}{n} \sum_{i=1}^{n} (P(t)_{i}-P(s)_{i})\\
 P(s)_{i} ={1}/{1+e^\frac{l_{c}}{T}
}
\label{eq:3}
 \end{aligned}
\end{equation}
where $P(t)_{i}$ and $P(s)_{i}$ are softened sigmoid predictions from teacher and student networks, respectively.
$l_{c}$ is the logit from the last fully connected layer in the network, and $T$ is the temperature for class $c$ \cite{liu2018multi}.
 
\subsection{Training}
We first train teacher network, which captures one view of context (e.g., global context) of HOIs along with hard labels, using cross-entropy loss. We then fix the teacher network and train the student network which is another view of HOIs (e.g., local context). Hence, the objective function for training the student network can be written as:
\begin{equation}
 L_{student} =\lambda_{1}L_{CE}+\lambda_{2} L_{Distill}
 \label{eq:4}
\end{equation}
where $L_{CE}$ is cross-entropy loss between student predictions and hard labels (e.g., ground truth). $\lambda_{1}$ and $\lambda_{2}$ are hyper-parameters for balancing the two losses and are set empirically as explained in Section \ref{subsubsection:as}.
For testing, the results is reported using only the student network.

\section{Experiments}
\subsection{Dataset and Settings}
\paragraph{Datasets.} 
We conduct intensive experiments on two public datasets, including Charades \cite{sigurdsson2016hollywood} and CAD-120 \cite{koppula2013learning} datasets. We choose these datasets because they include a variety of HOI categories. 

\begin{figure}[t]
  \centering
   \includegraphics[width=1.0\linewidth]{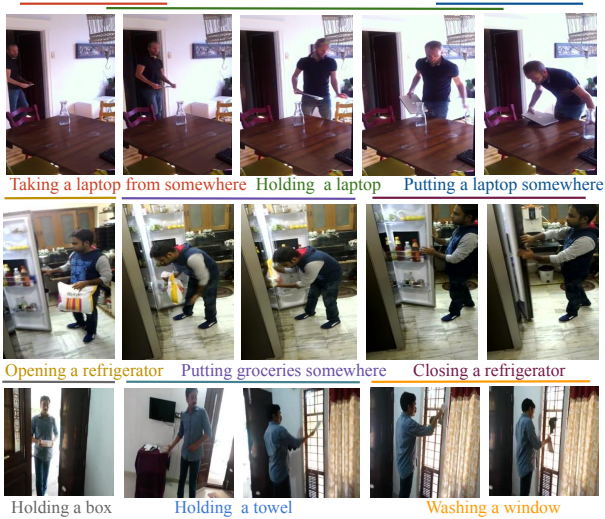}

   \caption{Examples of HOIs in videos from Charades dataset \cite{sigurdsson2016hollywood}.}
   \label{fig:char_ex}
\end{figure}
Charades dataset \cite{sigurdsson2016hollywood} consists of 9,848 multi-label videos with indoor daily activities that involve humans interacting with various types of objects. The number of videos in training phase is about 8K videos and 1.8K for validation. There are 157 action classes in total. Examples of some HOIs in Charades dataset are shown in Figure \ref{fig:char_ex}.

\begin{figure*}
  \centering
  \begin{subfigure}{0.17\linewidth}
     \includegraphics[width=1.0\linewidth]{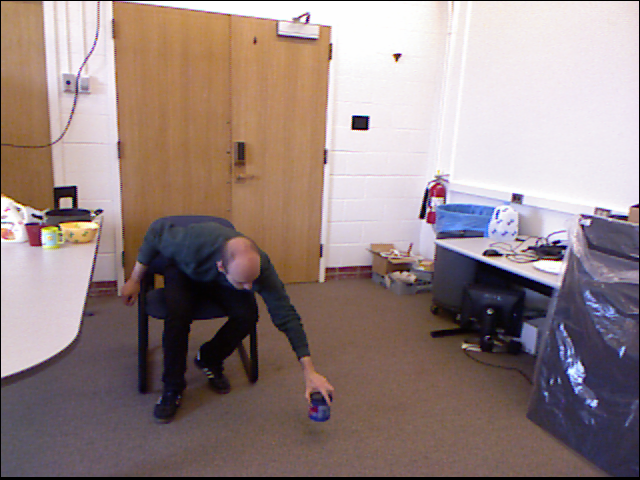}
    \caption{Picking objects}
    \label{fig:short-aa}
   \end{subfigure}
    \begin{subfigure}{0.17\linewidth}
     \includegraphics[width=1.0\linewidth]{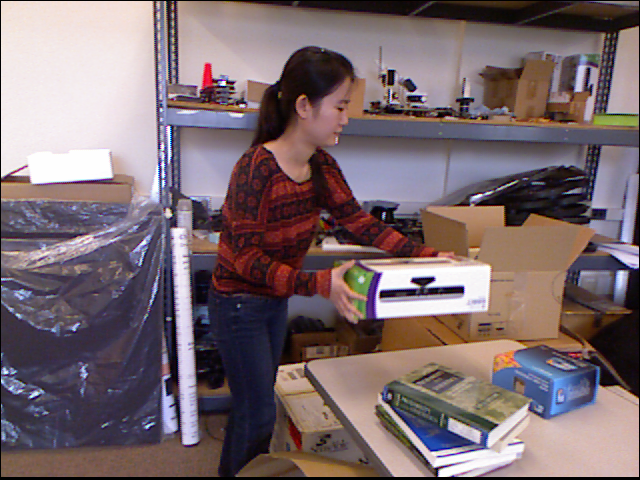}
    \caption{Arranging objects}
    \label{fig:short-bb}
   \end{subfigure}
   \begin{subfigure}{0.17\linewidth}
     \includegraphics[width=1.0\linewidth]{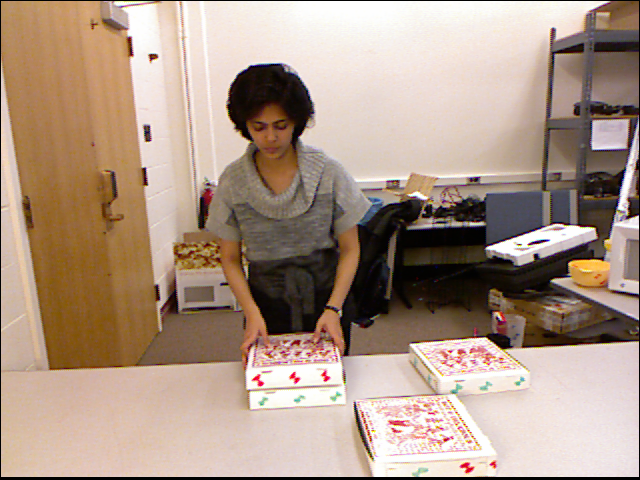}
    \caption{Stacking objects}
    \label{fig:short-c}
   \end{subfigure}
   \begin{subfigure}{0.17\linewidth}
     \includegraphics[width=1.0\linewidth]{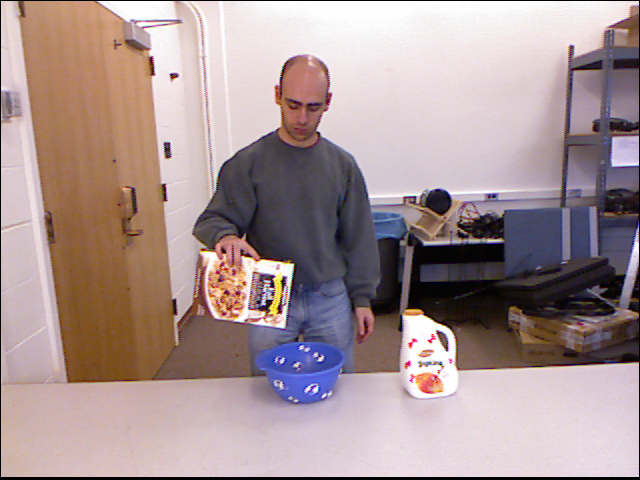}
    \caption{Making cereal }
    \label{fig:short-d}
   \end{subfigure}
   \begin{subfigure}{0.17\linewidth}
     \includegraphics[width=1.0\linewidth]{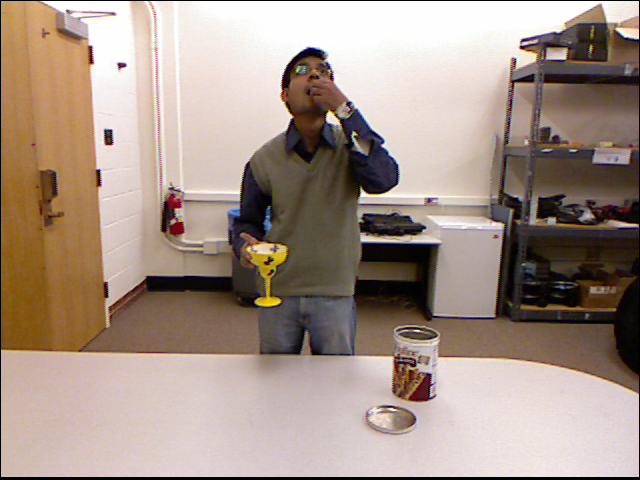}
    \caption{Taking medicine}
    \label{fig:short-e}
   \end{subfigure}
    \begin{subfigure}{0.17\linewidth}
     \includegraphics[width=1.0\linewidth]{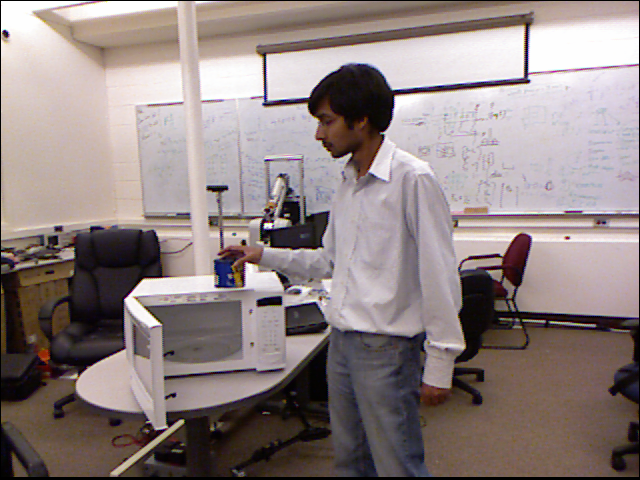}
    \caption{Taking food}
    \label{fig:short-f}
   \end{subfigure}
    \begin{subfigure}{0.17\linewidth}
     \includegraphics[width=1.0\linewidth]{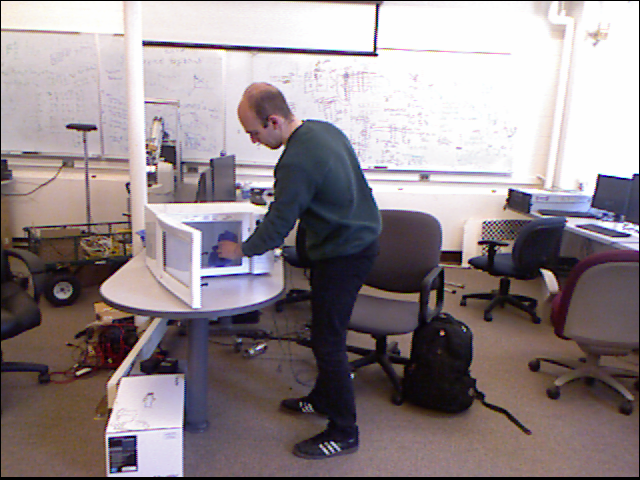}
    \caption{Cleaning objects}
    \label{fig:short-j}
   \end{subfigure}
   \begin{subfigure}{0.17\linewidth}
     \includegraphics[width=1.0\linewidth]{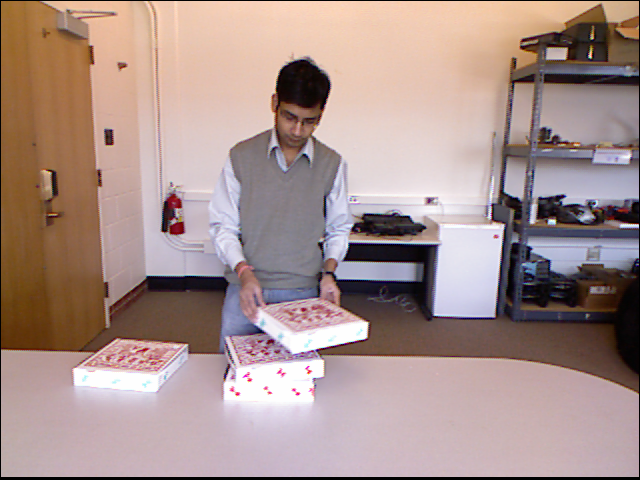}
    \caption{Unstacking objects}
        \label{fig:short-h}
   \end{subfigure}
   \begin{subfigure}{0.17\linewidth}
     \includegraphics[width=1.0\linewidth]{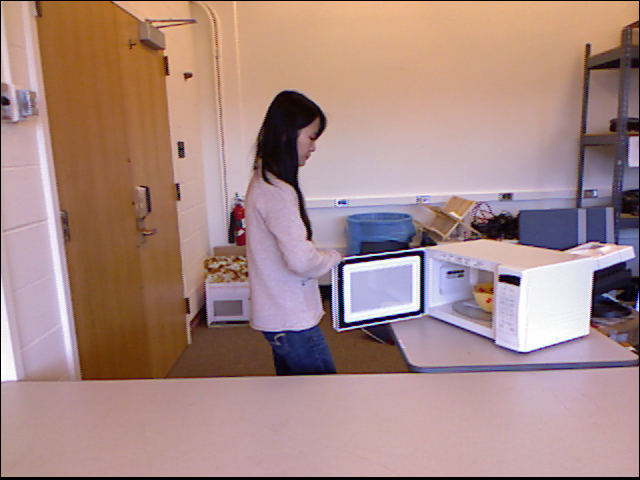}
    \caption{Microwaving food}
    \label{fig:short-k}
   \end{subfigure}
    \begin{subfigure}{0.17\linewidth}
     \includegraphics[width=1.0\linewidth]{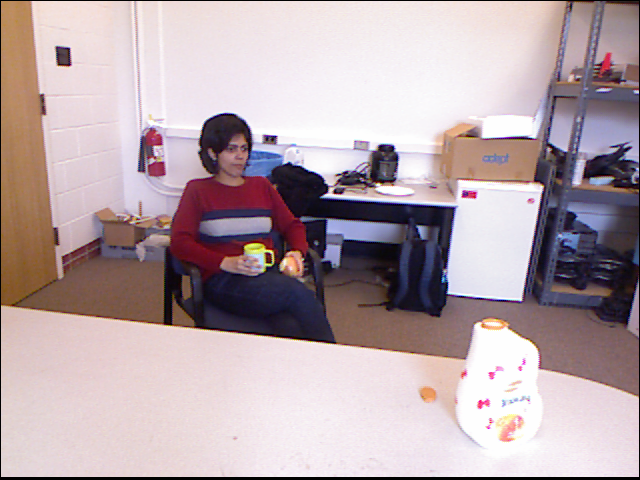}
    \caption{Having meal}
    \label{fig:short-l}
   \end{subfigure}
   \hfill
   
   \caption{Example of interaction activities in CAD-120 dataset \cite{koppula2013learning}.}
   \label{fig:short_all}
 \end{figure*}

Alternatively, CAD-120 \cite{koppula2013learning} contains 120 videos where 10 different daily life interactions are performed by 4 different subjects. Depth images and skeleton information are available besides RGB frames but we use only the RGB images. Figure \ref{fig:short_all} shows examples of these interactions. 
 
\paragraph{Evaluation Metric.} 
Since Charades dataset is a multi-label video dataset, we use mean average precision to report the final results. In contrast, each video in CAD-120 \cite{koppula2013learning} has only one activity label. Thus, accuracy is adopted as the evaluation metric as in \cite{sanou2019extensible}.

\subsection{Implementation Details}
\paragraph{Charades dataset.} For training our GLIDN, we follow training procedure in \cite{wang2018videos} and  we use Inflated 3D ConvNet (I3D) model \cite{carreira2017quo} with Resnet-50 and Slowfast-R50 \cite{feichtenhofer2019slowfast} as our backbone networks. In I3D backbone, we initialize it with pretrained parameters on Kinetics-400 dataset \cite{kay2017kinetics} from \cite{fan2020pyslowfast}.
For Slowfast-R50 backbone, we adapt it from \cite{fan2020pyslowfast} where it is already trained on Charades dataset. We sample 32 and 64 frames as in \cite{feichtenhofer2019slowfast} and \cite{wang2018videos})  from each video as input with 224×224 pixels for I3D and  Slowfast-R50, respectively. The inputs are randomly cropped such that the shorter side is sampled in [256, 320] pixels. We train I3D backbone for 60 epochs with a batch size of 8 videos, where the learning rate is set to 0.018 for the first 40 epochs and is reduced by a factor of 10 for the last 20 epochs. Following the previous works including \cite{wang2018videos,herzig2019spatio,tan2019object}, we use stage-wise training strategy where the model is trained end-to-end in the second stage for 30 epochs.

As in \cite{wang2018videos}, we apply RoIAlign on the output feature maps of the backbones (before the FC)  and each node in the graph is with a fixed dimension of 7$\times$7$\times$512 (1$\times$1$\times$512 via max pooling).

\muna{Since Charades dataset does not provide human and object bounding boxes, we use Region Proposal Network (RPN) in Faster R-CNN \cite{ren2015faster} to produce object proposals. We use the top 15 proposals at each frame. These proposal features (bounding boxes) represent human and object nodes in the graphs.}

We adapt binary cross-entropy with sigmoid activation as a loss function for multi-label video classification in addition to the distillation loss.

For inference, we perform multi-crop-view inference on each video. In other word, we sample 10 clips from each videos and perform multi-crop testing as in \cite{herzig2019spatio}. Later, the result is reported based on fusing scores from 30 views via max pooling.

\paragraph{CAD-120 dataset.} We sample 30 frames uniformally from each video and we used the bounding box annotations that are provided within the dataset. We follow \cite{sunkesula2020lighten} for extracting features for human and objects nodes. For each bounding box in a frame, we apply RoI cropping and then reshape it to meet the input size of 224$\times$224$\times$3 for 2D ResNet backbone. Therefore, human and object node features are with the size of 2048 dimension that are produced by ResNet-50.

Besides distillation loss, we train our model with the cross-entropy loss with an initial learning rate of 2.e-5. We train our model for 100 epochs in total using Adam optimizer \cite{kingma2014adam}. Our network is trained on a single Nvidia TITAN RTX 24GB GPU. \muna{Hyper-parameters for our training are summarized in Table \ref{table:reproductions}.}

\subsubsection{Comparison with State-of-the-Arts}
As shown in Table \ref{table:prior_cad} and Table \ref{table:prior_char}, we compare our GLIDN with all prior methods that applied on CAD-120 and Charades datasets, respectively. Our approach achieves the best performance. It is noted that on Charades, our network  outperforms the baselines including I3D and Slowfast, which do not consider spatio-temporal contextual views of objects.

\fred{Our network also performs better than STRG \cite{wang2018videos}, which has used spatio-temporal object relations. This implies that our approach of using different views of object relations via distillation can help the model generalize better in identifying different types of interactions.
More than this, our method has achieved better results even with much fewer number of proposals, as shown in Table \ref{table:proposals}.}

\fred{In addition, our approach of utilizing the two different views and their knowledge transfer can offer more informative cues about interaction even we do not use any human-object abstract information (e.g., the union of both objects) as in \cite{herzig2019spatio}.
This indicates the importance of context modeling of humans and objects without the need for additional information (e.g. visual phrases).}

Moreover, we noted that our choice of graph attention network for learning human-object relations in both global and local views is important since we achieved 35.35 comparing to 34.2 in\cite{wang2018videos} for the global context with fewer number of nodes.
Consequently, we have achieved the best results on Charades comparing to prior works that use the same backbone networks.

Furthermore, we have achieved better results on the CAD-120 \cite{koppula2013learning} than other works that use temporal sampling and 3D CNN \cite{sanou2019extensible,wang20143d} without fine tuning and with the use of \muna{object features extracted from 2D backbone}. 
\fred{This implies that our knowledge distillation from different views can notably contribute to HOIs reasoning, since it can better capture long-term temporal structure of interactions.}

\begin{figure}[t]
  \centering
   \includegraphics[width=1.2\linewidth]{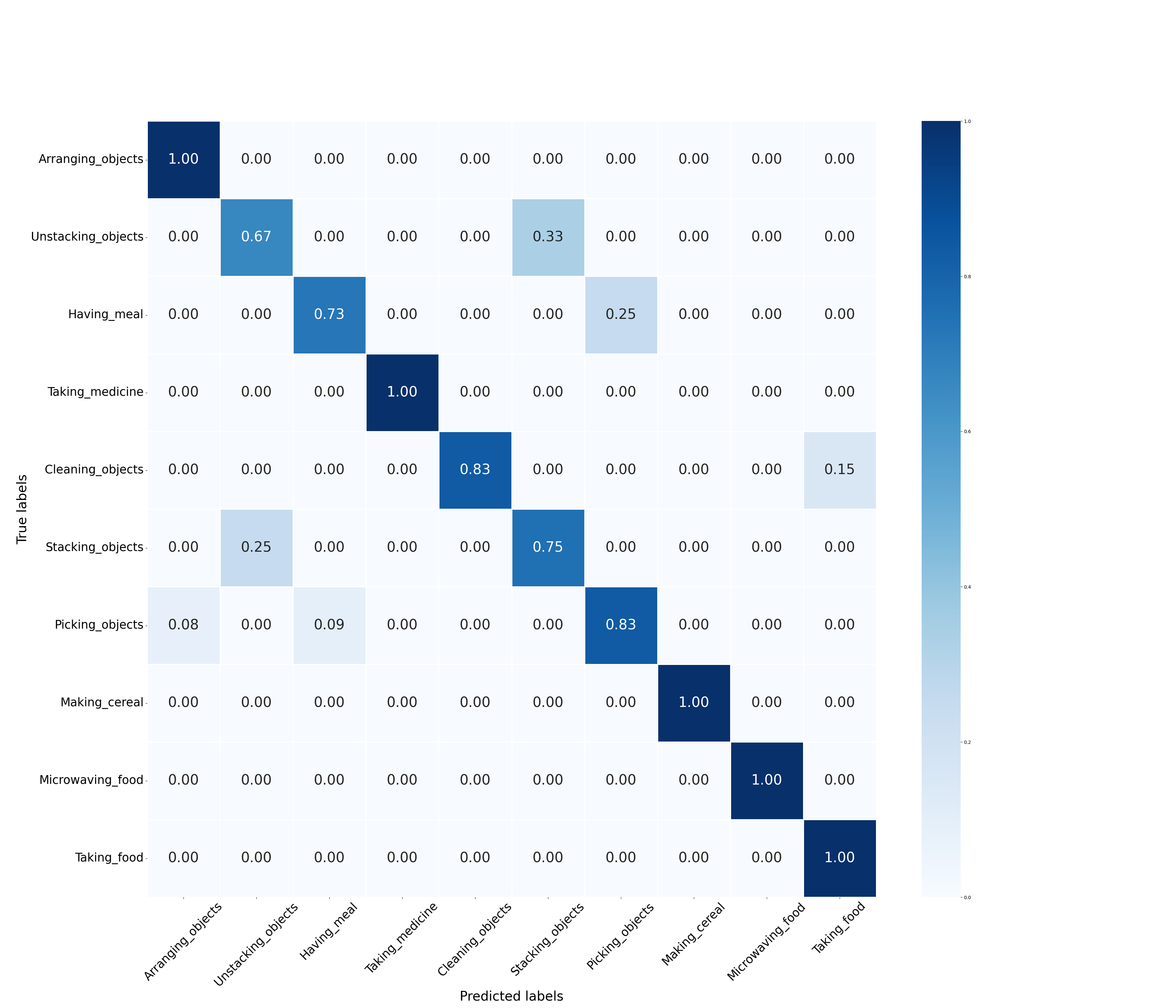}

   \caption{Confusion matrix for the CAD-120 dataset \cite{koppula2013learning} when using our proposed GLIDN.}
   \label{fig:cm}
\end{figure}

\fred{As shown in Figure \ref{fig:cm}, the confusion matrix studied how well our method can predict actions correctly based on CAD-120. It can be observed that most false predicted actions relate to stacking and unstacking objects or some actions alike. Such actions usually involve the same object but being different in human movement directions. This may be resolved by capturing more temporal information via increasing the number of sampled frames.}

\subsubsection{Ablation Studies}
\label{subsubsection:as}
To evaluate our proposed GLIDN, we conduct ablation studies to demonstrate the impact of each part of our GLIDN on learning HOIs. We first evaluate the baseline without any of interaction contextual views. We then evaluate our network by using each of the contextual views independently. Finally, we report the performance of our complete network. The ablation study results are shown in Table \ref{table:slowfast} and Table \ref{table:I3D} for Charades, while Table \ref{table:cad-120_abl} presents the results on CAD-120 dataset.
  \begin{table}
   \centering
   \begin{tabular}{@{}lccc@{}}
      \toprule
     Model & \# of nodes& Nodes info.& mAP\%\\
      \midrule
      STRG \cite{wang2018videos} &50&objects &36.20\\
      STRG \cite{wang2018videos} &25&objects &35.9\\
      STAG\cite{herzig2019spatio}&15&objects and edges*&37.20\\
      GLIDN \textbf{(ours)}&15&objects&\textbf{37.30}\\
    
      \bottomrule
   \end{tabular}
   \caption{Comparison of graph node settings with prior works on Charades \cite{sigurdsson2016hollywood}. 'Edges' means the union box of two object nodes. }
   \label{table:proposals}
  \end{table}
 
 \begin{table}
  \small
  
  \label{table:prior_Cad}
  \centering
  \begin{tabular}{lc}
    \toprule
    \cmidrule(r){1-2}
    Model &Accuracy\%      \\
    \midrule
    Wang et al. \cite{wang20143d} & 81.2\\
    *Liu et al.\cite{liu2020learning}&93.3\\
    *koppula et al.\cite{koppula2013learning}& 80.6\\
    *Tayyub et al. \cite{tayyub2014qualitative}&95.2\\
    Sanou et al. \cite{sanou2019extensible} & 86.4    \\ 
    GLIDN \textbf{(ours)} &\textbf{88.54}\\

    \bottomrule
  \end{tabular}
  \caption{Accuracy (\%) results on the CAD-120 dataset \cite{koppula2013learning}. '*' indicates that prior works make use of additional skeleton or depth information and thus are not directly comparable to our approach.}
  \label{table:prior_cad}
\end{table}

\begin{table}
  \centering
  \begin{tabular}{@{}lc@{}}
    \toprule
    Model & Accuracy\% \\
    \midrule
    Baseline & 74.17 \\
    Local-context (spatial) & 84.97\\
    Global-context (temporal) & 82.75 \\
    Local-teacher	& 87.7625 \\
    Global-teacher &	\textbf{88.54}\\
    \bottomrule
  \end{tabular}
  \caption{Ablation results on the CAD-120 dataset\cite{koppula2013learning}.}
  \label{table:cad-120_abl}
\end{table}

\begin{table*}
  \centering
  \begin{tabular}{@{}lcccccc@{}}
    \toprule
    Dataset & Optimizer&LR&Epochs&Decay&\# of GAT Layers & Training procedure \\
    \midrule
    CAD-120 \cite{koppula2013learning} & Adam& 2.e-5 &100&each 50 steps&3&  Leave-One-Out Cross-Validation\\
    Charades \cite{sigurdsson2016hollywood} & SGD&0.018&60,30& each 40 steps&1&Stage-Wise Training (2 stages)\\

    \bottomrule
  \end{tabular}
  \caption{A summary of training settings in our experiments on CAD-120 \cite{koppula2013learning}  and  Charades\cite{sigurdsson2016hollywood}.}
  \label{table:reproductions}
\end{table*}

\begin{table}
  \centering
  \begin{tabular}{@{}lc@{}}
    \toprule
    Model & mAP\%  \\
    \midrule
    Slowfast & 38.9 \\
    Local-context (spatial) & 40.73\\
    Global-context (temporal) & 39.95 \\
Local-teacher	& 39.89 \\
Global-teacher &\textbf{41.00}	\\
    \bottomrule
  \end{tabular}
  \caption{Ablation results on the Charades dataset \cite{sigurdsson2016hollywood} using Slowfast backbone.}
  \label{table:slowfast}
\end{table}

\begin{table}
  \centering
  \begin{tabular}{@{}lc@{}}
    \toprule
    Model & mAP\%  \\
    \midrule
    I3D & 34.23 \\
    Local-context (spatial) & 36.45\\
    Global-context (temporal) & 35.39 \\
    Global-teacher &36.81	\\
    Local-teacher	& \textbf{37.30} \\
    \bottomrule
  \end{tabular}
  \caption{Ablation results on the Charades dataset using I3D-R50 backbone.}
  \label{table:I3D}
\end{table}
\begin{table}
  \small

  \centering
  \begin{tabular}{llll}
    \toprule
    \cmidrule(r){1-2}
    Model &  Backbone &mAP\%      \\
    \midrule
    2-Stream \cite{sigurdsson2017asynchronous} & VGG-16  & 18.6       \\
    2-Stream +LSTM \cite{sigurdsson2017asynchronous} & VGG-16  & 17.8    \\
    Async-TF \cite{sigurdsson2017asynchronous} & VGG-16  & 22.4      \\
    a Multiscale TRN  \cite{zhou2018temporal} & Inception  & 25.2    \\
    I3D \cite{carreira2017quo} & Inception & 32.9    \\
    I3D \cite{wang2018videos} & R50-I3D & 31.8    \\
    STRG \cite{wang2018videos} & R50-I3D & 36.2    \\
    STAG \cite{herzig2019spatio} &R50-I3D  &  {37.2}\\
    Pose and Joint-Aware \cite{shah2020pose} &R50-I3D & 32.81\\
    GLIDN (ours) &R50-I3D  &  \textbf{37.30}\\
    LFB Max\cite{wu2019long} & R50-I3D-NL & 38.6\\
    Slowfast 16 x 8 \cite{feichtenhofer2019slowfast}& R50-3D    &38.9\\
    Slowfast 16 x 8$+$GLIDN \textbf{(ours)} & R50-3D   & \textbf{41.00}\\

    \bottomrule
  \end{tabular}
  \caption{Classification mAP (\%) results on the Charades dataset \cite{sigurdsson2016hollywood}.}
   \label{table:prior_char}
\end{table}

\paragraph{Are contextual views of humans and objects important?}
\muna{As shown in Tables \ref{table:cad-120_abl}, \ref{table:slowfast} and \ref{table:I3D}, 
\fred{running our network without any human-object relations or with only a single view (either local or global view) degrades the network performance.}}
It is clear that when we consider only human and object information (e.g., via concatenation) without learning their relation, the performance of the network decreases significantly by 14\% in CAD-120 \cite{koppula2013learning}. 

\muna{Also, when considering only human-object temporal relations on Charades, the performance drops by 1\% mAP, which reflects the importance of local relations between human and objects at a specific time as they can provide useful context information.} This indicates that some of the interactions can be recognized by focusing on the spatial relation especially with the existence of multiple objects around a human. Finally, capturing both the global and local human-object relations via distillation can help transfer the complementary information from the teacher view to the student contextual view. Hence, the ablation experiments illustrate that each component of the proposed GLIDN plays towards improving the the model performance where 41.00\% mAP is achieved on Charades.

\paragraph{Which of the contextual views play the roles of the teacher network?}
In the original form of the KD, the teacher network is larger than the student network. In contrast, in this work, the student and the teacher networks are both giving informative cues about interactions from different contextual views. 
\fred{Hence, we conduct comprehensive experiments to decide which of the contextual view can serve the role of the teacher. Logically, when we take into account the wide range of information provided by the global context, we can consider it as a larger view for HOIs since each human/object learned a relation with all other humans/objects throughout all video frames, while the local context only provides information about how humans/objects attend the others within each individual frame.}
This idea is evaluated on Charades \cite{sigurdsson2016hollywood} and CAD-120 datasets \cite{koppula2013learning}. As shown in Table \ref{table:cad-120_abl} and Table \ref{table:slowfast}, when we consider the global contextual view as the teacher, we achieved the best results. \fred{However, as shown in Table \ref{table:I3D}, when training Charades dataset with I3D backbone, we find that using the local contextual view as the teacher achieves better performance.}

This may be because in complicated HOI scenarios, the global contextual view comprises confusing HOIs, while individual local contextual view instead provides much clear interaction information. Also, because in slowfast experiments we use  objects from 64 frames which means that the temporal range of object information is wider comparing to the I3D backbone experiments where only humans and objects from 32 frames are used in constructing graph contexts. Hence, when the temporal range is not enough to capture better contextual information, especially in clutter background videos as in Charades \cite{sigurdsson2016hollywood}, \fred{the spatial local context teacher may outperform the temporal global one.}

Moreover, there are other factors that control the distillation which are the hyper-parameters of T (temperature), $\lambda_{1}$ and $\lambda_{2}$ (weights for balancing the losses in Eq. \ref{eq:4}). We conduct comprehensive experiments in both CAD-120 \cite{koppula2013learning} and Charades \cite{sigurdsson2016hollywood} using different values of these hyper-parameters. 
Two forms of $\lambda$ settings are used for balancing the weight between the two terms of the objective function as in Eq. \ref{eq:4}.
In the first form of setting, we used the generalized distillation form as in \cite{lopez2015unifying} where $\lambda_{1}$ is equal to (1-$\lambda_{2}$). The second form is by setting $\lambda_{1}$ to 1 and $\lambda_{2}$ to 4 as shown at the first row in Table \ref{table:hyper_meters} which shows the results of applying different hyper-parameters on CAD-120 dataset \cite{koppula2013learning} with different settings for teacher and student. 

We observed that the best values of T is different for both global contextual view and local view because each network view produces different probability distribution for the logits. Also, we find that in the global teacher, the temperature of 10 produces a good soft set of targets when the weight of the distillation loss is equal to 0.3 or 0.7. However, a higher value of T, such as 20, is better when the values of $\lambda_{1}$ and $\lambda_{2}$ are equal. Moreover, when we consider local contextual view as the teacher network, we find that small value of T (e.g., 5) with a distillation weight of 0.3 produces the best results of 87.76\%. Therefore, the optimal values of T and $\lambda$ can be set empirically based on the predictions of the teacher network.


\paragraph {Is teacher-student network design a good choice for distilling object contexts?}
In order to evaluate our teacher-student network design, we compare it with other collaborative learning approaches, such as Deep Mutual Learning (DML) \cite{zhang2018deep} where the two contexts views are jointly trained. \muna{As presented in Table \ref{table:dml}, we can observed that our teacher-student network achieves a better result of 88.54\% with an increase of 2\% when we consider the teacher network as the global context of HOIs where 86.64\% is achieved via DML.
This is because the teacher-student network approach allows the use of contextual information from the teacher network guiding the student network to capture much structural knowledge about HOIs.}

\begin{table}
  \centering
  \begin{tabular}{@{}lc@{}}
    \toprule
    Model & Accuracy\%  \\
    \midrule
    DML (local) &87.73	\\
    DML (global) & 86.64 \\
    our GLIDN (Local-teacher) & 87.76\\
    our GLIDN (Global-teacher) & \textbf{88.54} \\

    \bottomrule
  \end{tabular}
  \caption{Comparison between DML and teacher-student networks for distilling knowledge between object contexts. }
  \label{table:dml}
\end{table}

\begin{table}
  \small

  \centering
  \begin{tabular}{cccc}
    \toprule
    \cmidrule(r){1-2}
     T&$\lambda_{2}$ &Global-teacher\%& Local-teacher\%      \\
    \midrule
    2 & 4  &   87.56  & 84.36 \\
    5  & 0.3 &  88.36  & \textbf{87.76}\\
    10 & 0.3 &   88.45  & 83.53 \\
    20   & 0.3 &  87.62  &83.50\\
   5  & 0.5 &   84.33  &86.84\\
    10 & 0.5 &    85.69 &  84.25\\
    20   & 0.5 &  87.47  &83.59\\
    5  & 0.7 &    86.84&81.89\\
    10 & 0.7 &    \textbf{88.54} &82.61\\
    20   & 0.7 &  85.27  &86.00\\

    \bottomrule
  \end{tabular}
  \caption{Accuracy results on CAD-120 dataset \cite{koppula2013learning} after applying different values for weighting the distillation loss.}
   \label{table:hyper_meters}
\end{table}

\section{Conclusion}
The context of HOIs gives crucial cues about how human interacts with different objects. We propose GLIDN, a novel human objects interaction distillation network, which explicitly uses two different views of humans and objects context to capture their interactions at specific time and throughout a video. We also propose context knowledge distillation to transfer knowledge from the teacher contextual view of HOIs, to the student network that have information from different context of such interactions. Extensive experiments demonstrate the superiority of our approach over prior works on two datasets including Charades \cite{sigurdsson2016hollywood} and CAD-120 \cite{koppula2013learning}. As a future work, we will explore self-supervised approaches for identifying human and objects and their interactions in videos to overcome the need for human and object bounding boxes information, which are not available in most video datasets, while RPN may not accurately detect some objects.

{\small
\bibliographystyle{ieee_fullname}
\bibliography{PaperForReview}
}

\end{document}